\documentclass{article}

\PassOptionsToPackage{table}{xcolor}

\usepackage{microtype}
\usepackage{graphicx}
\usepackage{subcaption}
\usepackage{booktabs} 

\usepackage{hyperref}

\usepackage[accepted]{icml2026}

\usepackage{amsmath}
\usepackage{amssymb}
\usepackage{mathtools}
\usepackage{amsthm}

\usepackage{algorithm}
\usepackage{graphicx}

\usepackage{amsmath}
\usepackage{algorithmic}
\usepackage{textcomp}
\usepackage{subcaption}
\usepackage{tabularx}
\usepackage{color}
\usepackage{xcolor,soul}
\usepackage{multirow}
\usepackage{xspace}
\usepackage{adjustbox}
\usepackage{enumitem}
\usepackage{balance}
\usepackage{wrapfig, lipsum, booktabs}
\usepackage{colortbl}

\newdimen\abovecrulesep
\newdimen\belowcrulesep
\makeatletter
\patchcmd{\@@@cmidrule}{\aboverulesep}{\abovecrulesep}{}{}
\patchcmd{\@xcmidrule}{\belowrulesep}{\belowcrulesep}{}{}

\definecolor{demphcolor}{RGB}{144, 144, 144}
\definecolor{mygray}{gray}{0.4}
\definecolor{lightgray}{rgb}{0.9, 0.9, 0.9}

\usepackage{adjustbox}

\definecolor{tabhighlight}{HTML}{e5e5e5}
\definecolor{grey}{RGB}{128,138,135}

\definecolor{oorange}{RGB}{215,122,71}
\definecolor{yyellow}{RGB}{230,185,79}
\definecolor{ppurple}{RGB}{122,30,97}
\definecolor{ggreen}{RGB}{112,173,71}

\definecolor{battleshipgrey}{rgb}{0.3, 0.3, 0.3}
\definecolor{brilliantrose}{rgb}{1.0, 0.33, 0.64}
\definecolor{americanrose}{rgb}{1.0, 0.01, 0.24}
\definecolor{jweigreen}{rgb}{0,0.45,0.24}
\definecolor{bluegray}{rgb}{0.1, 0.1, 0.4}
\definecolor{ao(english)}{rgb}{0.0, 0.5, 0.0}
\definecolor{blanchedalmond}{rgb}{1.0, 0.92, 0.8}
\definecolor{atomictangerine}{rgb}{1.0, 0.6, 0.4}
\definecolor{chocolate(web)}{rgb}{0.82, 0.41, 0.12}
\definecolor{bananayellow}{rgb}{1.0, 0.88, 0.21}
\definecolor{goldenbrown}{rgb}{0.6, 0.4, 0.08}
\definecolor{aliceblue}{rgb}{0.94, 0.97, 1.0}
\definecolor{beige}{rgb}{0.96, 0.96, 0.86}
\definecolor{babyblue}{rgb}{0.54, 0.81, 0.94}
\definecolor{camel}{rgb}{0.76, 0.6, 0.42}
\definecolor{cinnamon}{rgb}{0.82, 0.41, 0.12}

\definecolor{redlinkcolor}{rgb}{0.79607843, 0.25098039, 0.25882353}
\definecolor{bluecitecolor}{rgb}{0,0.36,0.69}
\usepackage{hyperref}
\usepackage{booktabs}
\usepackage{makecell}
\usepackage{multirow}
\usepackage{colortbl}
\usepackage{adjustbox}
\usepackage{hyperref}
\usepackage[table]{xcolor}
\usepackage{tcolorbox}
\usepackage{subcaption}  
\usepackage{colortbl}
\usepackage{adjustbox}
\usepackage{makecell}
\usepackage{graphicx}
\usepackage{booktabs}
\usepackage[table]{xcolor}
\usepackage{xspace}
\usepackage{caption}

\usepackage{fontawesome5}  
\usepackage[bottom]{footmisc}  

\usepackage[capitalize,noabbrev]{cleveref}

\theoremstyle{plain}

\theoremstyle{definition}

\theoremstyle{remark}

\usepackage[textsize=tiny]{todonotes}

\newcommand{\task}{OmniDenseCaptioning\xspace}
\newcommand{\model}{TimeChat-Captioner\xspace}
\newcommand{\dataset}{TimeChatCap-42K\xspace}
\newcommand{\benchmark}{OmniDCBench\xspace}
\newcommand{\metric}{SodaM\xspace}

\icmltitlerunning{Scripting Multi-Scene Videos with Time-Aware and Structural Audio-Visual Captions}

\begin{document}

\twocolumn[
\icmltitle{TimeChat-Captioner: Scripting Multi-Scene Videos with Time-Aware \\ and Structural Audio-Visual Captions}



  \icmlsetsymbol{equal}{*}

  \begin{icmlauthorlist}
    \icmlauthor{Linli Yao}{pku}
    \icmlauthor{Yuancheng Wei}{scut}
    \icmlauthor{Yaojie Zhang}{estc}
    \icmlauthor{Lei Li}{hku}
    \icmlauthor{Xinlong Chen}{cas,comp}
    \icmlauthor{Feifan Song}{pku}
    \icmlauthor{Ziyue Wang}{pku}
    \icmlauthor{Kun Ouyang}{pku}
    \icmlauthor{Yuanxin Liu}{pku}
    \icmlauthor{Lingpeng Kong}{hku}
    \icmlauthor{Qi Liu}{hku}
    \icmlauthor{Pengfei Wan}{comp}
    \icmlauthor{Kun Gai}{comp}
    \icmlauthor{Yuanxing Zhang}{comp}
    \icmlauthor{Xu Sun}{pku}
  \end{icmlauthorlist}
  
  \icmlaffiliation{pku}{School of Computer Science, Peking University}
  \icmlaffiliation{scut}{South China University of Technology}
  \icmlaffiliation{estc}{University of Electronic Science and Technology of China}
  \icmlaffiliation{hku}{The University of Hong Kong}
  \icmlaffiliation{cas}{Institute of Automation, Chinese Academy of Sciences}
  \icmlaffiliation{comp}{Kling Team, Kuaishou Technology}

  \icmlcorrespondingauthor{Xu Sun}{xusun@pku.edu.cn}
  \icmlkeywords{Machine Learning, ICML}

\vskip 0.2in
\centerline{\includegraphics[width=1.0\textwidth]{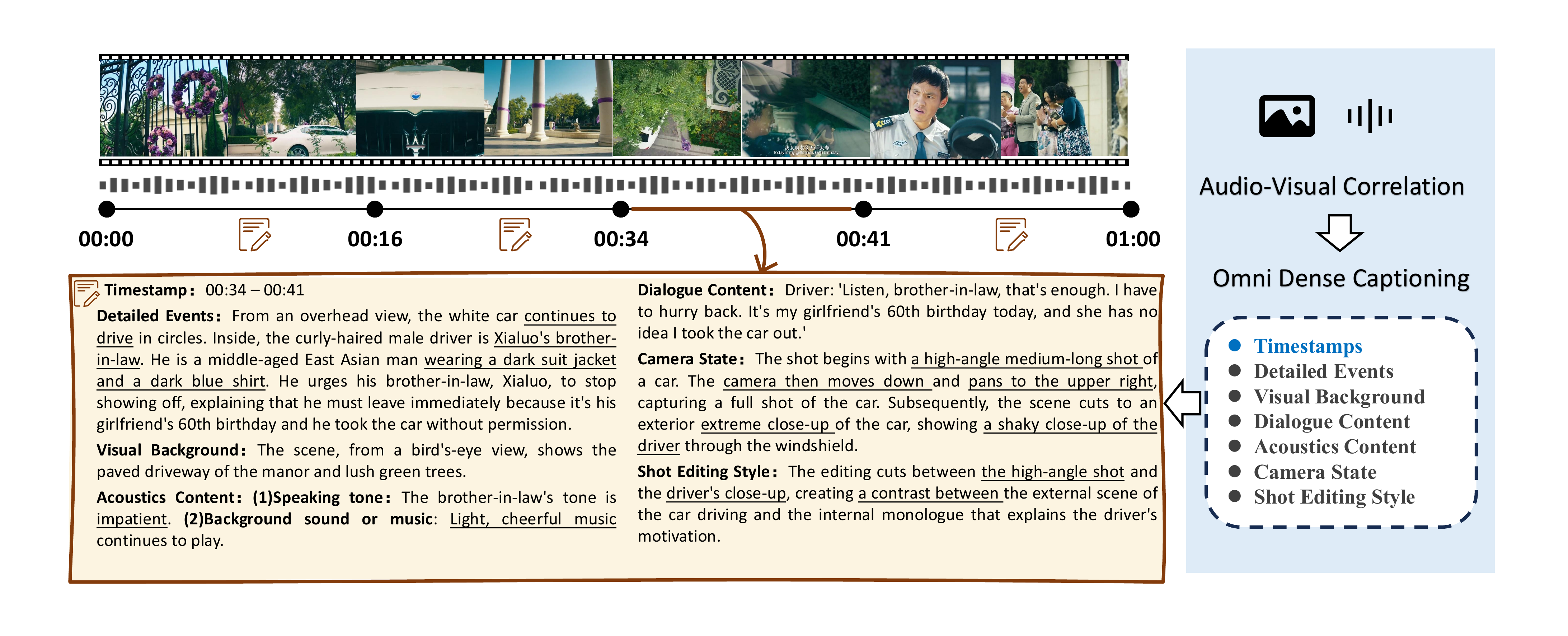}}
\vskip 0.1in
\captionof{figure}{\textbf{Illustration of the \task task.}
This paper introduces \textbf{Omni Dense Captioning} task, which generates fine-grained, temporally grounded descriptions for comprehensive audio-visual understanding. The term ``dense'' reflects two key properties: (1) \textbf{temporally-dense}: continuous scene segmentation with explicit timestamps, and (2) \textbf{description-dense}: structured captions spanning six dimensions: Audio-Visual Events, Visual Background, Camera State, Shot Editing, Dialogue, and Acoustic cues. These ``script-like'' descriptions allow readers to imagine the video scene-by-scene, as though reading a cinematic screenplay.}
\label{fig:teaser}

  \vskip 0.3in
]



\printAffiliationsAndNotice{Work done during Linli Yao's internship at Kling Team, Kuaishou Technology.}


\begin{abstract}
    This paper proposes Omni Dense Captioning, a novel task designed to generate continuous, fine-grained, and structured audio-visual narratives with explicit timestamps. To ensure dense semantic coverage, we introduce a six-dimensional structural schema to create ``script-like'' captions, enabling readers to vividly imagine the video content scene by scene, akin to a cinematographic screenplay. To facilitate research, we construct \benchmark, a high-quality, human-annotated benchmark, and propose \metric, a unified metric that evaluates time-aware detailed descriptions while mitigating scene boundary ambiguity. Furthermore, we construct a training dataset, \dataset, and present \model-7B, a strong baseline trained via SFT and GRPO with task-specific rewards. Extensive experiments demonstrate that \model achieves state-of-the-art performance, surpassing Gemini-2.5-Pro, while its generated dense descriptions significantly boost downstream capabilities in audio-visual reasoning (DailyOmni and WorldSense) and temporal grounding (Charades-STA). All datasets, models, and code are publicly available at \url{https://github.com/yaolinli/TimeChat-Captioner}.
\end{abstract}

\section{Introduction}
\label{sec:intro}

As video understanding~\cite{Qwen3-Omni,omnivinci,minicpm,video-SALMONN-o1,video-salmonn} and generation~\cite{omniavatar} enter the ``sound'' era, the alignment and interaction of omni-modal information (audio, visual, and text) have become pivotal research directions for Multimodal Large Language Models (MLLMs)~\cite{vita,Qwen3-VL,llava-video-sft,timechatonline}. 
Within this context, \text{Omni-Video Captioning}, which generates temporally grounded audio-visual-text triplets, emerges as a critical foundational task~\cite{videoSALMONN2,geng2025longvale,yuan2025tarsier2}.
Such high-quality audio-visual-text data provide comprehensive supervision signals that enable MLLMs to learn fine-grained cross-modal alignment during pre-training and post-training, while also benefiting downstream tasks such as Audio-Visual Reasoning~\cite{dailyomni,worldsense} and Video-to-Audio Generation~\cite{samaudio}.


However, a performant omni-video captioning framework, accompanied by a dedicated benchmark and evaluation suite, remains a largely unexplored frontier in the open-source community. \textbf{Existing audio-visual captioning}~\cite{videoSALMONN2,wu2025ugc} works primarily focus on generating global, paragraph-level descriptions without explicit timestamps. This lack of temporal granularity fails to provide the dense supervision signals necessary for MLLMs to master time-aware reasoning, such as temporal grounding~\cite{timer1}. On the other hand, \textbf{traditional dense video captioning} approaches~\cite{timechat,vid2seq} largely remain confined to the visual modality, neglecting the rich semantics embedded in audio. While recent advanced methods like LongVALE~\cite{geng2025longvale} have begun to incorporate audio cues, they predominantly focus on identifying salient events and generating concise summaries. This sparse and brief paradigm overlooks the continuous, fine-grained audio-visual nuances, thereby failing to capture the comprehensive semantics required for deep omni-modality alignment.

To bridge this gap, we propose a novel task \textbf{Omni Dense Captioning} with joint audio and visual semantics. Given a video with audio, the task goal is to semantically segment the input into continuous scenes and generate fine-grained audio-visual descriptions for each segment.
Specifically, ``dense'' here entails two aspects: \textbf{1) dense timestamps}, indicating continuous temporal segments that reveal the semantic scene changing and \textbf{2) dense captions}, referring to fine-grained descriptions covering the full audio-visual context (e.g., spatial attributes, actions, dialogue, and acoustic cues) along the temporal timeline. 
Unlike previous approaches that prioritize visual dominance, we explicitly enforce a six-dimension structural schema to ensure holistic audio-visual coverage:\textit{ (1) Overall Audio-Visual Events, (2) Background and Environment, (3) Camera State, (4) Multi-shot Editing Style, (5) Dialogue Content, and (6) Acoustic Cues}. This structured design aims to produce ``script-like'' data where reading the captions allows one to reconstruct the video in imagination, scene-by-scene. These  structural captions can serve as  abundant supervision signals and provide downstream MLLMs with sufficient context for omni-video understanding or generation.

To facilitate research in this direction, we construct a high-quality benchmark named \textbf{\benchmark}, comprising 1,122 human-annotated samples. Evaluating this task presents unique challenges, particularly the ambiguity of continuous scene boundaries. To address this, we propose \textbf{a novel unified metric \metric}, which jointly measures temporal timestamp accuracy and the semantic completeness of lengthy captions. \metric incorporates a dynamic programming alignment process to mitigate the time boundary gap between model predictions and human references. Finally, we present \textbf{a strong baseline \model-7B}, trained on synthesized high-quality data via Supervised Fine-Tuning (SFT) and Group Relative Policy Optimization (GRPO) stages. Extensive experiments demonstrate that \model not only achieves State-of-the-Art performance on \benchmark, surpassing Gemini-2.5-Pro~\cite{gemini}, but also generates rich semantics that boost performance on downstream Audio-Visual Reasoning tasks like Daily-Omni~\cite{dailyomni}, and WorldSense~\cite{worldsense}, and generalized to temporal grouding task Charades-STA~\cite{charades-sta}. We hope  \model will deliver dense temporal and textual supervision that significantly enhances MLLMs' omni-modal alignment capabilities.


\section{Related Work}
\label{sec:related_work}

\subsection{Audio-Visual Captioning}
Video captioning aims to generate textual descriptions of video content~\citep{wang2024tarsierrecipestrainingevaluating,yuan2025tarsier2advancinglargevisionlanguage}, with recent studies exploring fine-grained captioning that describes detailed temporal dynamics~\citep{zhong2025owlcap}. 
The emergence of omni-modal models~\citep{comanici2025gemini,qwen2d5omni,ai2025ming} has shifted research from vision-centric to joint audio-visual understanding~\citep{avocado}.
Representative works include AVoCaDO~\citep{avocado} for audiovisual temporal coherence, video-SALMONN-2~\citep{videoSALMONN2}, and UGC-VideoCaptioner~\citep{wu2025ugc} for multimodal integration.
However, these methods generate holistic captions without explicit temporal grounding. In contrast, \model outputs timestamped captions with structured, fine-grained descriptions for each scene.

\subsection{Time-Aware Video Captioning}
Dense video captioning~\citep{krishna2017dense} localizes temporal segments and generates event-level descriptions, evolving from pipeline-based to end-to-end frameworks~\citep{wang2021end,vid2seq,han2023shot2story20k}. 
Recently, LongVALE~\citep{geng2025longvale} advances long-range temporal modeling for extended video durations, while ARC-Chapter~\citep{pu2025arc} organizes videos into chapter-level units for structured descriptions.
Despite these advances, existing methods typically generate sparse, event-centric captions with concise outputs or focus only on salient events. In contrast, \task aims to capture comprehensive audiovisual semantics, producing multi-scene narratives with structured, fine-grained descriptions that cover all significant segments.

Beyond per-segment captioning, the broader multi-shot video literature spans both understanding~\citep{fang2024mmbenchvideo} and generation~\citep{an2025onestory}, both of which rely on dense, scene-aligned supervision that \task is designed to provide.

\subsection{Reinforcement Learning for Video Captioning}
Reinforcement learning (RL) ~\citep{schulman2017proximal,guo2025deepseek,zheng2025group,gao2025soft} has become an important paradigm in multimodal video understanding, particularly for aligning models with task-specific objectives~\citep{shao2025deepseekmath}.
CapRL~\citep{xing2025caprl} introduces verifiable rewards for caption generation, VideoCap-R1~\citep{meng2025videocap} incorporates structured reasoning steps, and AVoCaDO~\citep{avocado} extends GRPO~\citep{guo2025deepseek} with content coverage and length regularization rewards.
Unlike these approaches targeting holistic quality, we propose \metric, a reward that jointly optimizes temporal alignment and fine-grained coverage, applied within GRPO for temporally structured caption generation.

\begin{figure*}[t]
    \centering
    \includegraphics[width=\textwidth]{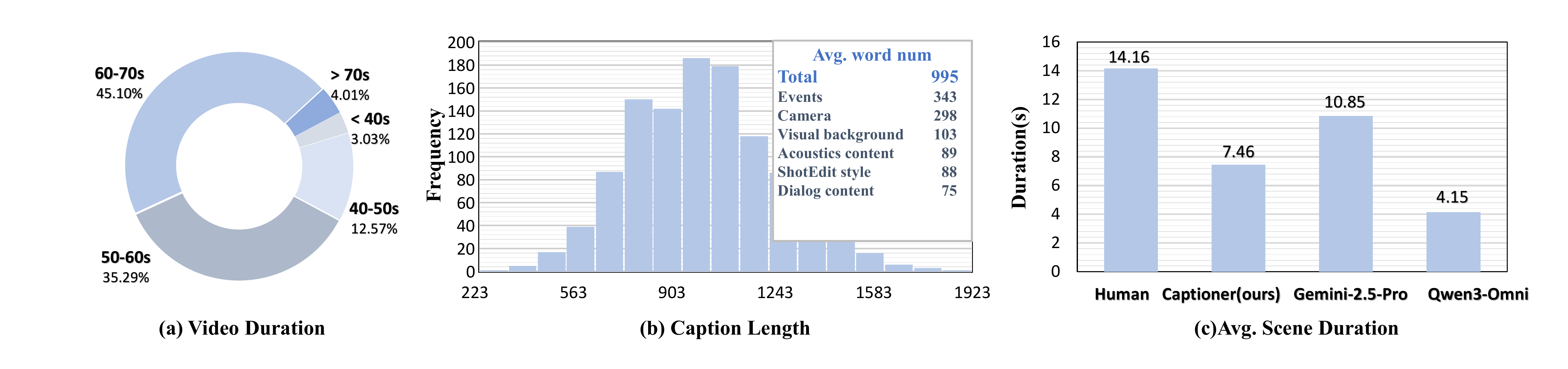}
    \caption{\textbf{Statistics of human-annotated \benchmark.} \textbf{(a)} Video duration distribution. \textbf{(b)} Caption length distribution with per-dimension. The benchmark features comprehensive annotations averaging 995 words per video. \textbf{(c)} Scene duration distribution (in seconds), compared against MLLM-generated outputs to highlight the granularity gap between human and model segmentations.}
    \label{fig:benchmark_stat}
\end{figure*}

\section{\task Task and A New Benchmark}
\label{sec:task_and_eval}

We first formally define the \task task (Section~\ref{subsec:task_definition}). We then introduce \benchmark, a high-quality benchmark with multi-dimensional scene-level annotations (Section~\ref{subsec:benchmark_dataset}). Finally, we propose \metric, a unified metric that jointly evaluates temporal segmentation and caption quality (Section~\ref{subsec:eval_metrics}).

\subsection{Task Definition}
\label{subsec:task_definition}
Given an input video $V$ with visual frames and audio signals, the goal of \task is to generate detailed paragraph-level descriptions with explicit timestamps that segment the video into successive multi-scenes. 
The description for each scene should cover comprehensively audio-visual details to achieve the goal of that by reading it, a user can imagine the scene-by-scene visual plots with synchronous audio information, as if they are watching the video.

\textbf{What is a ``Scene''?}
A scene is a semantically coherent video segment characterized by continuity in time, location, or narrative context. A single shot~\cite{han2023shot2story20k} refers to one continuous camera take. In contrast, a scene may consist of multiple shots that together convey a unified event or situation. Scene boundaries are usually indicated by clear transitions in visual setting, audio context, or narrative progression.

Formally, let the video $V$ be represented as a sequence of frames $F = \{f_1, f_2, \ldots, f_T\}$ and audio signals $A = \{a_1, a_2, \ldots, a_T\}$ over time steps $T$. The output script narrations $S$ can be expressed as a sequence of scene-level fine-grained captions $C = \{(t_1, c_1), (t_2, c_2), \ldots, (t_N, c_N)\}$, where each scene description $c_i$ encompasses structural and multiple-dimension audio-visual captions, and $t_i$ denotes the timestamp \texttt{MM:SS} indicating the start and end time of each scene $i$ in the video (e.g. ``\texttt{00:01-00:10}''). The number of scenes $N$ varies depending on the video's specific content.

Specifically, we design each scene description to comprehensively cover six dimensions: (1) \textit{Overall Audiovisual Events \textbf{(Events)}}: detailed narration of audiovisual content and actions; (2) \textit{Background and Environment \textbf{(Background)}}: depiction of the setting, location, and atmosphere; (3) \textit{Camera State \textbf{(Camera)}}: description of camera movements, angles, and framing; (4) \textit{Multi-shot Editing Style \textbf{(ShotEdit)}}: description of post-production editing techniques and how multiple shots are organized, such as montage sequences; (5) \textit{Dialogue Content \textbf{(Dialogue)}}: transcription and summary of spoken words and conversations with corresponding speakers; (6) \textit{Acoustic Cues \textbf{(Acoustic)}}: portrayal of background sounds, music, and auditory ambiance.

These dimensions collectively cover holistic spatial and temporal context, fine-grained visual-audio cues, camera state, and shot editing techniques to produce high-quality detailed descriptions. We highlight the critical differences between the \task task and existing dense video captioning~\cite{vid2seq,geng2025longvale} task:

\textbf{1) Comprehensive Visual-Audio Coverage}: Unlike dense video captioning that produces \textit{sparse} event descriptions focusing only on salient moments, \task aims to generate \textit{comprehensive} and \textit{successive} multi-scene narratives covering all significant scenes in a video, providing a holistic understanding of both visual and auditory content.

\textbf{2) Structured and Fine-grained Output}: Whereas existing methods typically provide brief descriptions spanning only a few sentences, \task is designed to generate \textit{structured}, comprehensive narratives across six distinct dimensions. This approach enables the capture of subtle, nuanced visual and audio details, resulting in richer and more informative captions.

Together, these two features make \task a comprehensive supervision source for both video \emph{understanding} and video \emph{generation} tasks. While prior annotation schemes target understanding alone (e.g., concise captioning, video QA, retrieval), our explicit modeling of \textit{Camera}, \textit{ShotEdit}, and \textit{Acoustic} dimensions further supplies the cinematic context required for downstream multi-scene or multi-shot video generation~\citep{an2025onestory}.

\subsection{Benchmark Dataset Curation}
\label{subsec:benchmark_dataset}

To support this novel and challenging task, we construct a high-quality benchmark \benchmark, through meticulous manual annotation.

\noindent\textbf{Data Source.} Ensuring video diversity and complexity is crucial for constructing representative multi-scene scripts. To this end, we curate a collection of high-resolution, clear-sound movie clips from Movie101~\cite{movie101v2}, as well as diverse general YouTube videos from YT-Temporal-1B~\cite{YT-Temporal-1B}, thus providing a broad range of content for our benchmark.

\noindent\textbf{Fully Manual Annotation Pipeline.} To ensure the highest quality and reliability of our benchmark, all data is carefully annotated and verified \emph{entirely by human experts} through a rigorous, systematic pipeline.

We structure the annotation process into three meticulous stages. \textbf{First}, crowd-sourced annotators review the pool of candidate videos, filtering out low-quality or unsuitable sources and assigning difficulty-level tags for annotation. \textbf{Second}, annotators watch each video in its entirety and segment it into multiple scenes, assigning continuous timestamps from a holistic perspective. \textbf{Third}, to annotate the six-dimensional scene descriptions, we assign different annotators to specific dimensions, as each requires distinct expertise. For instance, the \textit{Camera State} and \textit{Shot Editing Style} fields demand specialized knowledge of cinematography. 
To further ensure data integrity, both the timestamp and caption annotations are double-checked by independent annotators.

\noindent\textbf{Data Statistics.}
Through this rigorous annotation process, \benchmark comprises 1,122 videos with comprehensive and detailed multi-scene descriptions. As illustrated in Figure~\ref{fig:benchmark_stat}, a key characteristic of the dataset is the depth and richness of its annotations, with descriptions averaging 995 words per video.



\begin{figure*}[t]
    \centering
    \includegraphics[width=0.99\textwidth]{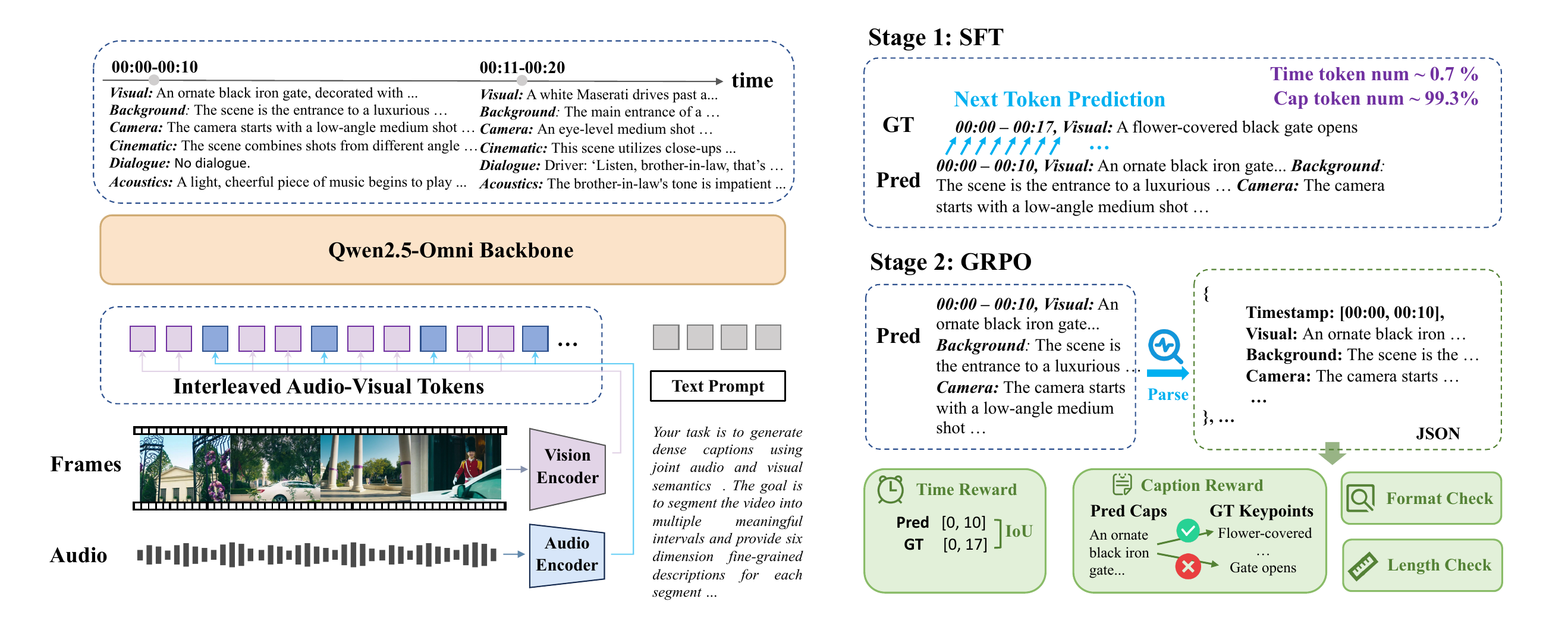}
    \caption{\textbf{Overview of \model Architecture.} \textbf{(Left)} This model leverages Qwen2.5-Omni~\cite{qwen2d5omni} with interleaved audio-visual tokens to generate multi-scene timestamps and six-dimensional captions. \textbf{(Right)} Two-stage training: SFT for task format learning, followed by GRPO with rewards for format, length, timestamp accuracy, and time-aware fine-grained caption quality.}
    \
    \label{fig:model_architecture}
\end{figure*}

\subsection{Evaluation Metric Design}
\label{subsec:eval_metrics}

An ideal evaluation framework for \task should measure the alignment between predicted and ground-truth outputs in terms of both temporal boundaries and descriptive content. This presents three key challenges: 
(1) assessing the accuracy of timestamp predictions (\textbf{Timestamp Accuracy}), 
(2) measuring the quality of fine-grained, multi-dimensional paragraph descriptions (\textbf{Caption Quality}), and 
(3) proposing a unified metric that jointly considers temporal alignment and caption quality across variable-length scene sequences (\textbf{Unified Metric}).

Formally, let the predicted output be $P = \{(\hat{t}_1, \hat{c}_1), \ldots, (\hat{t}_M, \hat{c}_M)\}$ and the ground-truth be $G = \{(t_1, c_1), \ldots, (t_N, c_N)\}$, where $M$ and $N$ denote the number of predicted and ground-truth scenes, respectively.

For clarity, we first outline the evaluation of timestamp accuracy and caption quality for a matched predicted and ground-truth scene pair $<(\hat{t}_i, \hat{c}_i), (t_j, c_j)>$.

\noindent\textbf{Timestamp Accuracy.} 
Given a predicted timestamp $\hat{t} = [\hat{t}_s, \hat{t}_e]$ and a ground-truth timestamp $t = [t_s, t_e]$, we compute the Intersection over Union (IoU)~\yrcite{dvc} as:

\begin{equation}
    \text{IoU}(\hat{t}, t) = \frac{|\hat{t} \cap t|}{|\hat{t} \cup t|}
\end{equation}

\noindent\textbf{Caption Quality.}
Conventional metrics (BLEU, METEOR, CIDEr) rely on n-gram matching and are ill-suited for paragraph-length, multi-dimensional descriptions. 
Drawing inspiration from recent advances~\cite{videoSALMONN2, avocado}, we employ the \textit{CheckList Score} for caption evaluation. Specifically, for each dimension $d \in \mathcal{D}$, the ground-truth caption $c$ is decomposed into a set of atomic elements $\mathcal{E}_d = \{e_1, e_2, \ldots, e_{|\mathcal{E}_d|}\}$. The predicted caption is then assessed against each of these elements as follows:
\begin{equation}
    \text{CheckList}(\hat{c}, c) = \frac{1}{\sum_{d \in \mathcal{D}} |\mathcal{E}_d|} 
    \sum_{d \in \mathcal{D}} \sum_{i=1}^{|\mathcal{E}_d|} \text{Judge}(\hat{c}, e_i)
\end{equation}
Here, $\text{Judge}(\hat{c}, e_i) \in \{0, 1\}$ indicates if $\hat{c}$ covers element $e_i$ from a judge model Gemini-2.5-Flash.
By averaging across all dimensions, we obtain the final CheckList score.



\noindent\textbf{A Unified Metric \metric.} 
The key challenge is jointly assessing timestamp accuracy and caption quality without a natural one-to-one correspondence between $M$ and $N$.
Since ``scene'' is an inherently semantic concept with ambiguous boundaries, different models (and even humans) may produce varying numbers of segments for the same video, as illustrated in Figure~\ref{fig:benchmark_stat} (c). This necessitates an alignment step before evaluation.


The core idea of \metric is a two-stage alignment strategy:
\begin{enumerate}[leftmargin=*]
    \item \textbf{IoU-based Dynamic Programming Alignment}: First, we find an optimal path through the $(M, N)$ scene grid of $\langle \text{pred}, \text{gt} \rangle$ pairs, using only temporal IoU as the scoring cost.
    \begin{equation}
        S[i][j] = \max\begin{cases}
            S[i-1][j] \\
            S[i][j-1] \\
            S[i-1][j-1] + \text{IoU}(t_i, \hat{t}_j)
        \end{cases}
    \end{equation}
    
    \item \textbf{Merging of Many-to-One Predictions}: Whenever multiple predicted scenes $\{\hat{p}_k, \ldots, \hat{p}_{k+l}\}$ are aligned to the same ground-truth scene $g_i$, we concatenate their captions and expand their timestamp range to form a single merged prediction:
    \begin{align}
        \hat{t}_{\text{merged}} &= [\min(\hat{t}_{k,s}, \ldots), \max(\hat{t}_{k,e}, \ldots)] \\
        \hat{c}_{\text{merged}} &= \text{Concat}(\hat{c}_k, \ldots, \hat{c}_{k+l})
    \end{align}
    where $s$ and $e$ denote the start time and end time of each predicted scene, respectively.
    This handles the common case where MLLMs generate finer-grained (shorter) segments than the ground-truth as Figure~\ref{fig:benchmark_stat} (c) shown. We only merge predictions while keeping the ground-truth unchanged to ensure evaluation fairness.
\end{enumerate}

After alignment, we obtain $K$ temporally matched pairs $\mathcal{M} = \{(\hat{p}_1, g_1), \ldots, (\hat{p}_K, g_K)\}$ where $K \leq N$. We then compute Timestamp Accuracy for each matched $\langle \text{pred}, \text{gt} \rangle$ pair and report the F1 score across thresholds $\{0.3, 0.5, 0.7, 0.9\}$, as well as the mean IoU, to assess overall segmentation quality following~\cite{dvc}. To evaluate temporally-aware caption quality, we calculate the CheckList Score for each matched pair and compute the F1 score for all pairs to obtain the final \metric score following~\cite{soda}.

\noindent\textbf{Summary.} Compared to $\text{SODA}_\text{c}$~\cite{soda}, \metric: (1) reduces judge-model cost from $O(MN)$ to $O(K)$ where $K \leq N$ by decoupling IoU matching from text evaluation, and (2) gracefully handles many-to-one alignments through merging, mitigating scene boundary ambiguity while ensuring holistic semantic coverage.
Extensive human evaluation in Appendix~\ref{sec:human_eval} confirms that \metric correlates well with human judgment.


\definecolor{rowgrayL}{gray}{0.93} 
\definecolor{rowgrayD}{gray}{0.89}
\definecolor{lightline}{gray}{0.85} 

\begin{table*}[t]
\centering
\caption{\textbf{Quantitative comparison on the \task task.} \textbf{Bold} and \underline{underline} highlight the best and second-best results among open-source models, respectively. $^{\dagger}$ indicates expert models specialized in temporal-aware captioning. \metric is the primary metric reflecting the quality of temporally-aligned, multi-dimensional captions.}
\label{tab:script_table}
\small
\renewcommand{\arraystretch}{1.3}
\setlength{\tabcolsep}{5pt} 

\resizebox{\textwidth}{!}{
\begin{tabular}{lc|ccccccccc}
\toprule
\multirow{2.5}{*}{\textbf{Model}} & \multirow{2.5}{*}{\textbf{Modality}} & \multicolumn{2}{c}{\textbf{Multi-Scene Seg.}} & \multicolumn{7}{c}{\textbf{Time-aware Dense Captioning Quality}} \\ 
\cmidrule(lr){3-4} \cmidrule(lr){5-11}
 & & \textbf{F1} & \textbf{mIoU} & \textbf{Camera} & \textbf{Events} & \textbf{Background} & \textbf{Acoustics} & \textbf{ShotEdit} & \textbf{Dialogue} & \makecell{\textbf{\metric} \\ \textbf{(Avg.)}} \\ 
\midrule

\rowcolor{white} \multicolumn{11}{c}{\textbf{Proprietary Models}} \\ \midrule
Gemini-2.5-Pro & V + A & 68.5 & 74.9 & 8.1 & 48.1 & 39.1 & 25.4 & 34.5 & 46.4 & 33.7 \\
Gemini-2.5-Flash & V + A & 45.6 & 53.1 & 11.5 & 38.1 & 42.1 & 22.4 & 27.6 & 42.6 & 30.0 \\ 
\midrule

\rowcolor{white} \multicolumn{11}{c}{\textbf{Open-source Models}} \\ \midrule
LongVALE$^{\dagger}$(7B)~\yrcite{geng2025longvale} & V + A & 45.2 & 55.6 & 0.8 & 0.6 & 1.3 & 0.5 & 3.1 & 5.0 & 1.8 \\ 
Qwen2.5-Omni(7B)~\yrcite{qwen2d5omni} & V + A & 37.2 & 43.4 & 1.6 & 3.9 & 12.3 & 3.4 & 3.3 & 15.7 & 4.6 \\
MiniCPM-o-2.6(8B)~\yrcite{minicpm} & V + A & 49.9 & 60.2 & 1.2 & 5.3 & 16.5 & 1.5 & 7.4 & 11.3 & 5.4 \\
OmniVinci(9B)~\yrcite{omnivinci} & V + A & 29.0 & 39.7 & 1.6 & 8.2 & 15.7 & 1.3 & 6.7 & 14.3 & 6.9 \\
Qwen3-Omni(30B-A3B)~\yrcite{Qwen3-Omni} & V + A & 54.8 & 64.2 & 3.1 & 20.2 & 21.6 & 5.1 & 14.1 & 25.4 & 14.3 \\
\arrayrulecolor{lightline}\hline\arrayrulecolor{black} 

\rowcolor{rowgrayL} \textbf{\model-7B (SFT)} & \textbf{V + A} & \textbf{62.4} & \textbf{70.8} & \underline{8.9} & \underline{30.5} & \underline{36.3} & \underline{30.6} & \underline{33.9} & \underline{44.8} & \underline{32.6} \\
\rowcolor{rowgrayD} \textbf{\model-7B (GRPO)} & \textbf{V + A} & \underline{61.2} & \underline{69.6} & \textbf{12.4} & \textbf{39.6} & \textbf{49.2} & \textbf{38.2} & \textbf{43.5} & \textbf{54.3} & \textbf{35.0} \\ 
\bottomrule
\end{tabular}
}
\end{table*}

\section{\model Framework}

We introduce \model, a specialized Video Large Language Model tailored for the \task task. Built upon joint audio-visual understanding, \model achieves accurate multi-scene timestamp prediction while generating fine-grained, structured descriptions for each segment.

\subsection{Overall Architecture}

As illustrated in Figure~\ref{fig:model_architecture}, we build \model upon the Qwen2.5-Omni~\cite{qwen2d5omni} backbone, leveraging its Thinker module for joint audio-visual perception with the Vision Encoder from Qwen2.5-VL~\cite{qwen2.5vl} and the Audio Encoder from Qwen2-Audio~\cite{qwen2-audio}. 

This backbone incorporates two key designs tailored to the requirements of \task. First, it arranges audio and visual tokens in a temporally interleaved sequence, enabling synchronous cross-modal comprehension—unlike traditional methods that process each modality in isolation~\cite{geng2025longvale}. Second, it employs Multimodal Rotary Position Embedding (M-RoPE)~\cite{Qwen2-VL} to encode absolute temporal positions, thereby facilitating precise scene boundary localization and continuous timestamp prediction.

\subsection{Training Data Collection}

To construct high-quality training data for \task, we develop a synthetic data pipeline powered by Gemini-2.5-Pro, as depicted in Figure~\ref{fig:data_pipeline}. This pipeline proceeds through three stages: video source selection, a two-step caption generation process, and  quality filtering.

\noindent\textbf{Video Source Sampling.}
We curate videos from two complementary datasets: (1) MMTrail-2M~\yrcite{chi2024mmtrail}, which is donimant and features a diverse and carefully cleaned collection of trailer videos spanning a broad range of topics; and (2) Movie101~\yrcite{movie101v2}, which consists of movie commentary videos with abundant and rich audiovisual content. 
To balance annotation quality and information density, we segment the raw videos into 3-minute clips.

\noindent\textbf{Two-Step Construction Pipeline.}
Recognizing that Gemini-2.5-Pro cannot reliably produce high-quality task data in a single pass, we adopt a coarse-to-fine approach:
\begin{itemize}[leftmargin=*]
    \item \textit{Stage 1: Boundary Segmentation.} Gemini-2.5-Pro analyzes each 3-minute clip to generate temporal segmentations accompanied by brief captions (e.g., ``0:00-0:15: a boy singing...'').
    \item \textit{Stage 2: Detailed Caption Generation.} Using the Stage 1 segmentations as scaffolding, Gemini-2.5-Pro is prompted to produce fine-grained, multi-dimensional descriptions for each segment, comprehensively covering all six dimensions outlined in Section~\ref{subsec:task_definition}. Detailed prompts are provided in the appendix.
\end{itemize}

\noindent\textbf{Data Quality Filtering.}
We ensure the fidelity of the training data through careful filtering:
videos with fewer than two scene segments, lacking audio tracks, containing JSON formatting errors or missing caption fields, as well as segments below a minimum duration, are all excluded from the final dataset.

\noindent\textbf{Summary.} Upon completion of filtering, we obtain 42K high-quality time-aware video-caption pairs, which constitute the final training dataset.
It is worth noting that our training data is entirely independent from the benchmark in terms of both video sources and annotation schema (synthetic annotations for training vs. manual annotations for evaluation), ensuring a fair assessment of generalization.

\subsection{Training Strategy}
\label{subsec:training}

\task is a challenging task that requires both accurate temporal segmentation and lengthy, structured textual output. To build a performant specialist model, we adopt Supervised Fine-Tuning (SFT) to teach the model the task format, followed by Group Relative Policy Optimization (GRPO)~\cite{shao2025deepseekmath} strategy to jointly improve timestamp accuracy and caption quality.


\definecolor{rowgrayD}{gray}{0.89}
\definecolor{catgray}{gray}{0.94}
\definecolor{lightline}{gray}{0.85}

\begin{table}[t]
\centering
\caption{\textbf{Caption-based Results on Omni-VideoQA Benchmarks.} Captions generated by each model are fed to Gemini-2.5-Pro to answer QA questions, so higher accuracy reflects richer and more complete captions.}
\label{tab:videoqa}
\small
\renewcommand{\arraystretch}{1.2}
\setlength{\tabcolsep}{4pt}

\resizebox{\columnwidth}{!}{
\begin{tabular}{l c|cc}
\toprule
\textbf{Model} & \textbf{Size} & \textbf{Daily-Omni} & \textbf{World-Sense} \\
\midrule

\rowcolor{catgray} \multicolumn{4}{l}{\textit{\,\,Closed-source Models}} \\
Gemini-2.5-Pro~\yrcite{gemini}                 & --       & 60.2 & 33.8 \\
Gemini-2.5-Flash~\yrcite{gemini}               & --       & 55.3 & 31.0 \\
\midrule

\rowcolor{catgray} \multicolumn{4}{l}{\textit{\,\,Open-source Models}} \\
HumanOmniV2~\yrcite{yang2025humanomniv2}       & 7B       &  8.2 &  6.6 \\
ARC-Hunyuan-Video~\yrcite{ge2025arc}           & 7B       &  8.6 &  8.7 \\
MiniCPM-o-2.6~\yrcite{minicpm}                 & 8B       &  9.8 &  7.2 \\
Qwen2.5-Omni~\yrcite{qwen2d5omni}              & 7B       & 13.4 &  8.6 \\
UGC-VideoCaptioner~\yrcite{wu2025ugc}          & 3B       & 17.0 & 11.2 \\
video-SALMONN-2~\yrcite{videoSALMONN2}         & 7B       & 29.9 & 18.2 \\
Qwen3-Omni-Instruct                            & 30B-A3B  & 17.5 & 12.7 \\
Qwen3-Omni-Captioner                           & 30B-A3B  & 27.2 & 14.1 \\

\arrayrulecolor{lightline}\hline\arrayrulecolor{black}

\rowcolor{rowgrayD} \textbf{Ours (GRPO)} & \textbf{7B} & \textbf{52.8} & \textbf{22.6} \\
\bottomrule
\end{tabular}
}
\end{table}

\begin{figure}[t]
\centering
\includegraphics[width=0.95\columnwidth]{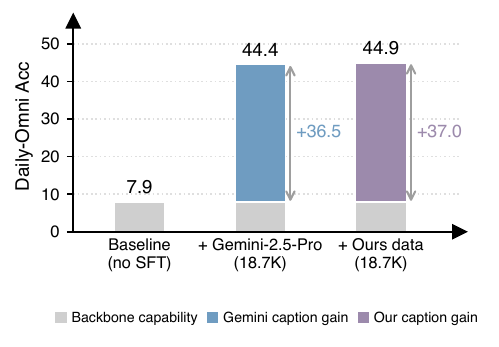}
\caption{\textbf{Caption Quality as Training Data.} We fine-tune Qwen2.5-Omni-3B via LoRA~\cite{lora}  on 18.7K video-caption samples annotated by two captioners and report the resulting caption-based Daily-Omni accuracy (the same caption-based VideoQA protocol as Table~\ref{tab:videoqa}). Captions from our open-source \model match the supervision quality of the closed-source Gemini-2.5-Pro.}
\label{fig:trainutility}
\end{figure}

\subsubsection{SFT Stage}
We first fine-tune the Qwen2.5-Omni backbone on our  training data using standard next-token prediction loss~\cite{sft-survey}. The input consists of raw video frames and audio wavs, while the target output follows our structured format with timestamps and multi-dimensional captions. This stage enables the model to follow the basic output format and preliminarily learn this complex task.

\subsubsection{GRPO Stage}
While SFT teaches the model to mimic the training distribution, it has inherent limitations for \task:
\begin{itemize}[leftmargin=*]
    \item \textbf{Token Imbalance}: Timestamp-related tokens constitute only a small fraction of the output (0.7\%), while caption tokens dominate. Standard cross-entropy loss treats all tokens equally, providing insufficient gradient signal for accurate temporal prediction.
    \item \textbf{Limited Generalization}: SFT models tend to overfit to the scene count distribution in training data, struggling to generalize to videos with varying numbers of scenes.
\end{itemize}

To address these issues, we adopt Group Relative Policy Optimization (GRPO)~\yrcite{shao2025deepseekmath}, a reinforcement learning algorithm that eliminates the need for a separate critic model as in PPO. For each training sample with input $q$, we sample $G$ candidate outputs $\{o_1, o_2, \ldots, o_G\}$ from the current policy $\pi_{\theta_{\text{old}}}$ and compute their rewards $\{r_1, r_2, \ldots, r_G\}$. The advantage for each response $o_i$ is computed relative to the group:
\begin{equation}
    A_i = \frac{r_i - \text{mean}(\{r_1, \ldots, r_G\})}{\text{std}(\{r_1, \ldots, r_G\})}
\end{equation}
The policy is then optimized using the following objective:

\small

\begin{align}
    &\mathcal{J}_{\text{GRPO}}(\theta) = \mathbb{E} \Bigg[ \frac{1}{G} \sum_{i=1}^{G} \min \bigg( \frac{\pi_\theta(o_i|q)}{\pi_{\theta_{\text{old}}}(o_i|q)} A_i, & \notag \\ 
    & \text{clip}\Big(\frac{\pi_\theta(o_i|q)}{\pi_{\theta_{\text{old}}}(o_i|q)}, 1-\epsilon, 1+\epsilon \Big) A_i \bigg) - \beta \cdot \mathbb{D}_{\text{KL}}\big(\pi_\theta \| \pi_{\text{ref}}\big) \Bigg] &
\end{align}

\normalsize

where $\epsilon$ is the clipping threshold and $\beta$ controls the KL divergence penalty from a reference policy $\pi_{\text{ref}}$.

\noindent\textbf{Reward Design.} We design task-specific rewards to boost timestamp accuracy and caption quality:
\begin{itemize}[leftmargin=*]
    \item \textit{Format Reward} $\mathcal{R}_F$: A binary reward indicating whether the output can be parsed as a valid JSON list $P = \{(\hat{t}_1, \hat{c}_1), \ldots, (\hat{t}_M, \hat{c}_M)\}$. If parsing succeeds, $\mathcal{R}_F = 1$; otherwise $\mathcal{R}_F = 0$.
    
    \item \textit{Length Reward} $\mathcal{R}_L$: To prevent the model from generating overly lengthy outputs prone to hallucination or repetitive content that fails to terminate, we apply a length-regularized reward following~\cite{avocado}
    \item \textit{Timestamp Reward} $\mathcal{R}_T$: The average F1 score at IoU thresholds $\{0.3, 0.5, 0.7, 0.9\}$ between predicted and groundtruth timestamps along the optimal alignment path (as introduced in Section~\ref{subsec:eval_metrics}).
    
    \item \textit{Time-aware Caption Reward} $\mathcal{R}_C$: We adopt the unified \metric metric as the reward to encourage comprehensive and temporally-aligned structural captions.
\end{itemize}


\definecolor{rowgrayD}{gray}{0.9} 
\definecolor{lightline}{gray}{0.85}

\begin{table}[t]
\centering
\caption{\textbf{Temporal grounding performance on Charades-STA.} All models are fine-tuned on the Charades-STA training set. Models marked with $\dagger$ are expert models. } 
\label{tab:grounding}
\small
\renewcommand{\arraystretch}{1.3}
\setlength{\tabcolsep}{8pt}

\resizebox{\columnwidth}{!}{
\begin{tabular}{l|cccc}
\toprule
\multirow{2.5}{*}{\textbf{Method}} & \multicolumn{4}{c}{\textbf{Charades-STA}} \\ 
\cmidrule(lr){2-5}
 & \textbf{R1@0.3} & \textbf{R1@0.5} & \textbf{R1@0.7} & \textbf{mIoU} \\ 
\midrule

TimeChat$^{\dagger}$~\yrcite{timechat} & -- & 46.7 & 23.7 & -- \\
TimeSuite$^{\dagger}$~\yrcite{timesuite} & 79.4 & 67.1 & 43.0 & -- \\
TimeExpert$^{\dagger}$~\yrcite{timeexpert} & -- & 64.1 & 43.3 & -- \\ 

\arrayrulecolor{lightline}\midrule\arrayrulecolor{black} 

Qwen2.5-Omni-7B & 78.3 & 65.9 & 44.1 & 56.7 \\ 

\rowcolor{rowgrayD} \textbf{Ours} & \textbf{79.8} & \textbf{68.7} & \textbf{48.3} & \textbf{58.8} \\ 
\bottomrule
\end{tabular}
}
\end{table}

The final reward $\mathcal{R}$ is a weighted sum of these components:
\begin{equation}
    \mathcal{R} = \alpha_f \cdot R_F + \alpha_l \cdot R_L + \alpha_t \cdot R_T + \alpha_c \cdot R_C
    \label{eq:reward}
\end{equation}
where each hyperparameters $\alpha$ controls the contribution of its respective reward. 

\section{Experiments}
\label{sec:experiments}

\subsection{Experimental Setup}
We adopt a two-stage training pipeline: SFT on 40K training samples for 2 epochs (lr=5e-5, batch size=128), followed by GRPO on 2K training samples for 1 epoch (lr=1e-5, batch size=64, rollout=8). Reward weights for format, length, temporal, and caption quality are set to 0.5, 0.5, 1.0, and 1.0, respectively. Videos are sampled at 2 FPS and the training maximum sequence length is 32K tokens. All experiments are conducted on 32$\times$80\textsc{G} GPUs. Further details are provided in the Appendix.




\definecolor{rowgrayD}{gray}{0.9}

\begin{table}[t]
\centering
\small
\caption{\textbf{Ablation study on training data scale and reward components.}
\metric denotes the time-aware caption reward $R_C$.
Both SFT data scaling and GRPO contribute to the final performance.
In particular, adding the \metric reward in GRPO further improves caption quality.
}
\label{tab:ablation}
\begin{tabular}{l|cc}
\toprule
\textbf{Model Variant} & \textbf{\benchmark} & \textbf{Daily-Omni} \\
\midrule
Qwen2.5-Omni & 4.6 & 13.4 \\
\midrule
+ SFT (20K) & 31.3 & 49.3 \\
+ SFT (40K) & 32.6 & 50.7 \\
\quad + GRPO (w/o \metric) & 32.5 & 50.4 \\
\rowcolor{rowgrayD} \textbf{\quad + GRPO (w/ \metric)} & \textbf{35.0} & \textbf{52.8} \\
\bottomrule
\end{tabular}
\end{table}

\subsection{Main Results on \benchmark}
As summarized in Table \ref{tab:script_table}, \model achieves highly competitive results on the Omni-Video Scripting Benchmark. Regarding scene boundary localization, our model ranks second only to the industry-leading proprietary model, Gemini-2.5-Pro, while significantly outperforming all other open-source baselines. For time-aware captioning quality, evaluated via the \text{SodaM} metric (an aggregate average across six dimensions: \textit{camera}, \textit{events}, \textit{background}, \textit{acoustics}, \textit{shot editing}, and \textit{dialogue}), \model-GRPO achieves state-of-the-art performance with a score of \text{35.0}. This result even surpasses the strongest closed-source model, Gemini-2.5-Pro, demonstrating that RL effectively improves accurate scene segmentation and fine-grained captioning.  Quantitative cases are visualized in Figure~\ref{fig:case}.


\begin{table*}[t]
\centering
\caption{\textbf{Effect of Dynamic Programming (DP) Merging on \metric.} We report \metric scores with and without the DP merging design (detailed in Section~\ref{subsec:eval_metrics}), along with the head-to-head human Elo win rate between the two comparison models. The ground-truth (GT) average segment duration of scenes is 14.2\, seconds.
}
\label{tab:dp_merging}
\small
\renewcommand{\arraystretch}{1.15}
\setlength{\tabcolsep}{8pt}
\begin{tabular}{l|c|cc|c}
\toprule
\textbf{Model} & \textbf{Avg. Duration (Pred / GT)} & \textbf{\metric (w/ DP)} & \textbf{\metric (w/o DP)} & \textbf{Human Win Rate} \\
\midrule
Qwen2.5-Omni (7B)           & 7.0\,s\, / \,14.2\,s & \phantom{0}4.6 & 4.4 & 16.7\% \\
Qwen3-Omni (30B-A3B)        & 4.2\,s\, / \,14.2\,s & 14.3           & 5.6 & 75.0\% \\
\midrule
Result Gap ($\Delta$) & --- & \text{+9.7} & +1.2 & +58.3 \\
\bottomrule
\end{tabular}
\end{table*}

\subsection{Results on Omni-VideoQA Benchmarks}
To assess whether \model's captions support downstream audio-visual reasoning, we adopt a caption-based VideoQA protocol: each model first generates a video description, which is then fed to Gemini-2.5-Pro as the sole context for QA. The resulting accuracy thus serves as a proxy for caption completeness and fidelity.

As summarized in Table~\ref{tab:videoqa}, although \model is optimized primarily for temporal-aware dense captioning, it exhibits strong cross-task transfer to general audio-visual captioning, yielding better caption-based VideoQA results: on \textit{Daily-Omni} and \textit{World-Sense}, it attains 52.8 and 22.6, respectively, outperforming all open-source baselines by a clear margin. This suggests that the temporally-dense, multi-dimensional audio-visual semantics encoded in our captions provide richer downstream supervision than conventional video captioning targets, even under the distributional shift between training and evaluation video domains.

\subsection{Results on Temporal Grounding Benchmarks}
To evaluate the generalization and transferability of our model in fine-grained temporal video understanding, we report the fine-tuning results on the Charades-STA~\yrcite{charades-sta} benchmark in Table \ref{tab:grounding}. \text{\model-GRPO} achieves superior performance across all evaluation metrics. Remarkably, our model consistently outperforms established expert models specifically designed for temporal video understanding tasks, such as TimeSuite and TimeExpert, as well as the Qwen2.5-Omni-7B baseline. These results validate that trained on \task task with \dataset significantly enhances the model's fundamental temporal understanding, thereby strengthening performance on downstream temporal grounding tasks.

\subsection{Caption Quality as Training Data}
\label{subsec:training_utility}
Beyond benchmarking, \model can serve as a cost-effective, open-source data engine for training Omni-VideoLLMs. To validate this, we annotate 18.7K additional videos with \model-GRPO and use the resulting captions to fine-tune Qwen2.5-Omni-3B with LoRA~\cite{lora}. For a fair comparison, we also fine-tune the same backbone on 18.7K captions produced by the closed-source Gemini-2.5-Pro.
As shown in Figure~\ref{fig:trainutility}, captions from \model yield downstream performance on Daily-Omni comparable to those from Gemini-2.5-Pro (44.9 vs.\ 44.4 in accuracy), while the off-the-shelf baseline reaches only 7.9. This demonstrates that \model is a practical open-source alternative to closed-source captioning APIs for scaling dense audio-visual supervision.

\subsection{Ablation Studies}


\noindent\textbf{Impact of Data Scale.}
We first evaluate the effect of supervised SFT data quantity in Table \ref{tab:ablation}. Increasing the training data from 20K to 40K samples leads to a consistent performance gain across all benchmarks, with the \benchmark score rising from 31.3 to 32.6. This demonstrates that more scripting data provides a stronger performance.

\noindent\textbf{Effectiveness of \metric Reward.}
Base rewards (format, length, and temporal alignment) ensure structural validity and basic temporal accuracy. Our ablation focuses on the unified \metric reward ($R_{C}$), which targets time-aware caption quality. As shown in Table~\ref{tab:ablation}, removing $R_{C}$ yields a stable but limited baseline, while incorporating $R_{C}$ significantly improves both temporal understanding and caption completeness, and even yields remarkable gains on the out-of-domain DailyOmni benchmark. This validates that optimizing for time-aware dense captioning quality serves as an effective proxy task for enhancing general audiovisual comprehension capabilities. Notably, the GRPO strategy with merely 2K training samples proves more effective than scaling up SFT training data from 20K to 40K, demonstrating the efficiency of our reward-guided optimization approach.

\noindent\textbf{Effectiveness of DP Merging.}
\label{subsec:dp_merging_main}
We validate the DP many-to-one merging step (Section~\ref{subsec:eval_metrics}) on two models with independently established human preference: humans prefer Qwen3-Omni (30B-A3B) over Qwen2.5-Omni (7B) in \textbf{75.0\%} of pairwise comparisons. As shown in Table~\ref{tab:dp_merging}, both models over-segment relative to the ground-truth (14.2\,s), but Qwen2.5-Omni happens to predict longer segments (7.0\,s) than Qwen3-Omni (4.2\,s). Without DP merging, hard one-to-one matching unfairly rewards this incidental granularity match: the \metric gap between the two models collapses to a misleading $+1.2$, sharply contradicting the 75\% human preference. With DP merging, the gap widens to $+9.7$, faithfully reflecting human judgment. This confirms that DP merging prevents \metric from being skewed by segment granularity rather than caption quality. Setup and additional safeguards are deferred to Appendix~\ref{subsec:dp_merging}.




\section{Conclusions}
\label{sec:conclusion}

We introduce \task, a novel task for generating temporally-aligned, multi-dimensional, and structurally rich video captions. We present the high-quality, human-annotated \benchmark benchmark and propose tailored metrics such as \metric to advance research in this area. Our specialized \model model, trained with synthetic audio-visual data and task-specific rewards, outperforms the proprietary Gemini-2.5-Pro and demonstrates strong generalization to related omni-video understanding tasks. Beyond benchmarking, we further demonstrate that \model can serve as a cost-effective, open-source captioning data engine that matches closed-source APIs in supervising downstream Omni-VideoLLMs. We hope our approach will promote comprehensive omni-video understanding and support future multi-scene video generation by providing fine-grained data.

\section*{Acknowledgements}

This research was partially supported by the National Natural Science Foundation of China under Grant No.~92470205. Xu Sun is the corresponding author.

\section*{Impact Statement}

This paper advances omni-video understanding through dense, temporally-grounded audiovisual captioning. Positive impacts include improved accessibility for impaired users and enhanced video-based education. Potential risks involve inherited biases from pre-trained models and possible misuse for misinformation. 

To ensure transparency and mitigate risks: (1) all videos in our training and evaluation datasets are sourced exclusively from publicly available academic datasets (MMTrail~\cite{chi2024mmtrail} and Movie101~\cite{movie101v2}), with no private data collected; (2) we document all data sources and model limitations; and (3) we release resources under responsible use licenses. We believe the benefits outweigh the risks when appropriate safeguards are followed.
\nocite{langley00}

\balance
\bibliography{example_paper}
\bibliographystyle{icml2026}

\newpage
\appendix
\onecolumn



\begin{figure*}[h!]
    \centering
    \includegraphics[width=\textwidth]{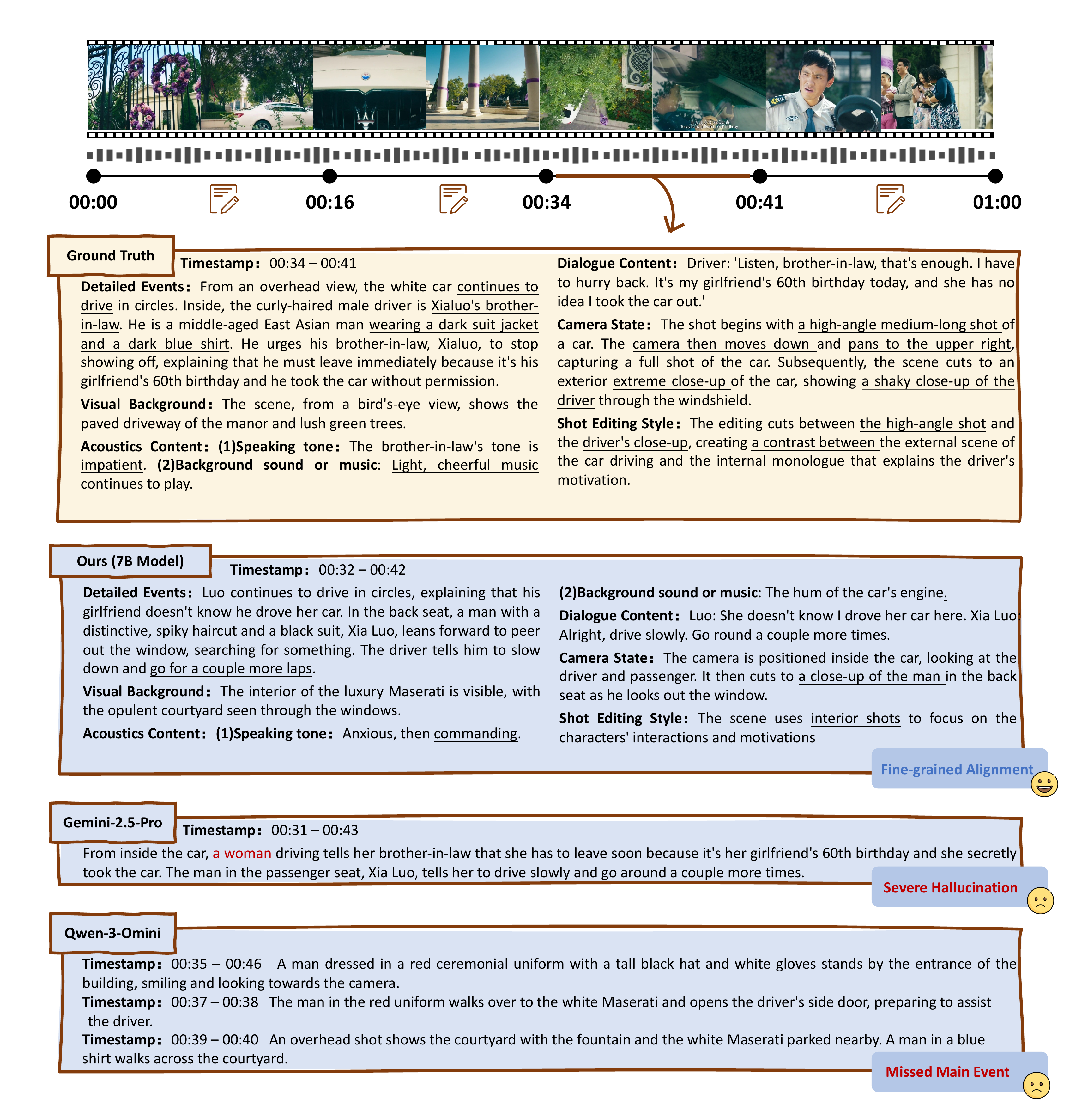}
    \caption{\textbf{Qualitative case analysis.} We compare \model~with Gemini-2.5-Pro and Qwen-3-Omni on a sample from \benchmark. Our model achieves \underline{\textit{fine-grained alignment}} with the ground truth across all six annotation dimensions: detailed events, visual background, acoustics, dialogue, camera state, and shot editing style. In contrast, Gemini-2.5-Pro~\cite{gemini} exhibits severe hallucination by misidentifying the male driver as a woman, fundamentally distorting the scene semantics. Qwen-3-Omni~\cite{Qwen3-Omni} \textit{misses the main event} entirely, describing irrelevant background elements (a doorman in red uniform) while ignoring the central conversation inside the car. These results demonstrate \model's superior capability in accurate character recognition, faithful event grounding, and comprehensive multi-dimensional annotation.}
    \label{fig:case}
\end{figure*}

\section{Limitations and Future Work.}
\label{sec:limitations}
Our current work has several limitations that warrant future investigation. 
\textbf{First}, the 32K context-window constraint poses a significant challenge during training. Since the \task task involves both lengthy inputs (video frames sampled at 2 FPS) and outputs (captions averaging 1K words), extending the context window is essential to accommodate more video frames and generate comprehensive captions. 
\textbf{Second}, our model demonstrates limited generalization to videos of varying durations, particularly hour-long content. To address this, we currently adopt a segment-then-caption strategy: dividing long videos into shorter clips (approximately one minute each) and applying \model sequentially to generate fine-grained omni-captions for each segment.

In future work, we plan to address these limitations through two directions: (1) collecting more diverse long-form videos to improve duration generalization and timestamp segmentation accuracy, and (2) incorporating efficient techniques such as token compression~\cite{yao2026survey} to reduce the sequence length of audio-video-text inputs, thereby lowering training costs, especially during the GRPO stage.


\section{Human Elo Evaluation and \metric Validation}
\label{sec:human_eval}

To rigorously validate \metric beyond automatic statistics, we conduct three complementary studies that examine: (i)~the \emph{robustness} of \metric across different LLM judges (Section~\ref{subsec:multi_judge}); (ii)~the \emph{human alignment} of \metric compared with traditional captioning metrics (Section~\ref{subsec:human_alignment}); and (iii)~the \emph{necessity} of the dynamic-programming (DP) segment-merging step inside \metric (Section~\ref{subsec:dp_merging}). The human study follows the blind pairwise Elo protocol introduced by AuroraCap~\cite{chai2024auroracap}.

\subsection{Multi-Judge Cross-Validation}
\label{subsec:multi_judge}

\noindent\textbf{Motivation.}
\metric scores rely on an LLM judge to assess the quality of long, multi-dimensional captions. A natural concern is whether the metric inherits the bias of any single judge---does the model ranking change if we swap the judge? To probe this, we re-compute \metric using four independent LLM judges spanning both proprietary models (Gemini-2.5-Flash, GPT-5.1, Claude-3.5-Haiku) and an open-source model (DeepSeek-V3.2).

\noindent\textbf{Setup.}
We re-evaluate the same five systems on \benchmark with each of the four judges in turn, keeping the prompt template and DP-merging procedure fixed. We measure cross-judge agreement using Kendall's coefficient of concordance $W$ over the four induced rankings ($W{=}1$ denotes perfect agreement).

\begin{table}[h]
\centering
\caption{\textbf{\metric scores under four different LLM judges.} Although absolute scores naturally differ across judges, the \emph{relative ranking} of models is preserved: our GRPO model is the top-1 system under every judge, surpassing the closed-source Gemini-2.5-Pro. Kendall's coefficient of concordance $W{=}0.925$ indicates very strong agreement.}
\label{tab:multi_judge}
\small
\renewcommand{\arraystretch}{1.15}
\setlength{\tabcolsep}{8pt}
\begin{tabular}{l|cccc}
\toprule
\textbf{Model} & \textbf{Gemini-2.5-Flash} & \textbf{GPT-5.1} & \textbf{DeepSeek-V3.2} & \textbf{Claude-3.5-Haiku} \\
\midrule
Qwen2.5-Omni-7B          &  4.6 &  5.9 & 23.8 & 12.7 \\
Qwen3-Omni-30B-A3B       & 14.3 & 14.7 & 19.8 & 23.0 \\
Ours (SFT)               & 32.6 & 31.5 & 44.5 & 58.1 \\
Gemini-2.5-Pro           & 33.7 & 33.7 & 44.7 & 49.0 \\
\rowcolor{rowgrayD} \textbf{Ours (GRPO)} & \textbf{35.0} & \textbf{34.4} & \textbf{48.6} & \textbf{62.4} \\
\bottomrule
\end{tabular}
\end{table}

\noindent\textbf{Findings.}
As Table~\ref{tab:multi_judge} shows, while absolute scores vary across judges (e.g., Claude-3.5-Haiku tends to assign higher absolute values), the \emph{relative} ordering of the five systems is highly consistent. Our GRPO model ranks first under every judge, surpassing Gemini-2.5-Pro. The high concordance ($W{=}0.925$) confirms that \metric's ranking is largely judge-agnostic, mitigating concerns of single-judge bias.

\subsection{Human Alignment: \metric vs.\ Traditional Captioning Metrics}
\label{subsec:human_alignment}

\noindent\textbf{Motivation.}
Robustness across judges is necessary but not sufficient: a metric must also \emph{agree with human preferences}. Traditional captioning metrics (CIDEr, METEOR, SODA\_c) are known to correlate weakly with human judgment in lengthy, multi-dimensional captioning settings, where surface-level $n$-gram overlap fails to capture cross-modal semantic completeness. We therefore directly compare the human alignment of \metric against these standard baselines.

\noindent\textbf{Setup.}
Following the Elo protocol of AuroraCap~\cite{chai2024auroracap}, three human experts conducted \textbf{129 blind pairwise A/B comparisons} over the five systems on \benchmark (anonymized output, randomized order, $\sim$13 comparisons per ordered model pair). After removing tied judgments, we obtain $N{=}122$ non-tie comparisons. We report two complementary measures:
\begin{itemize}[leftmargin=1.6em, nosep, topsep=2pt]
    \item \textbf{Case-Level Agreement} (primary, $N{=}122$): for each non-tie pair, whether the per-video metric ordering matches the human preference. This directly tests whether the metric makes the same fine-grained judgements as humans.
    \item \textbf{Pearson $r$} (supportive, $N{=}5$ systems): the system-level correlation between metric scores and Elo ratings derived from human judgments. Because $N{=}5$ is small, we treat this as supportive rather than definitive.
\end{itemize}

\begin{table}[h]
\centering
\caption{\textbf{Human alignment of \metric vs.\ traditional captioning metrics.} \metric achieves $\sim$78\% case-level agreement with human pairwise preferences---substantially higher than CIDEr, METEOR, and SODA\_c---and the agreement is stable across all four LLM judges. Model-level Pearson $r$ is reported as a supportive measure (computed over 5 systems).}
\label{tab:human_alignment}
\small
\renewcommand{\arraystretch}{1.15}
\setlength{\tabcolsep}{10pt}
\begin{tabular}{l|ccc}
\toprule
\textbf{Metric} & \textbf{Case-Level Agree.} ($\uparrow$) & \textbf{Pearson $r$} ($\uparrow$) & \textbf{$p$-value} \\
\midrule
CIDEr     & 47.5\% & 0.437 & 0.462 \\
METEOR    & 55.7\% & 0.167 & 0.789 \\
SODA\_c   & 60.7\% & 0.553 & 0.334 \\
\midrule
\metric (DeepSeek-V3.2)    & 70.5\% & 0.933 & 0.021 \\
\metric (GPT-5.1)          & 76.2\% & 0.954 & 0.012 \\
\rowcolor{rowgrayL} \metric (Gemini-2.5-Flash) & \textbf{77.9\%} & \textbf{0.960} & \textbf{0.010} \\
\rowcolor{rowgrayL} \metric (Claude-3.5-Haiku) & \textbf{77.9\%} & \textbf{0.960} & \textbf{0.010} \\
\bottomrule
\end{tabular}
\end{table}

\noindent\textbf{Findings.}
As Table~\ref{tab:human_alignment} shows, \metric reaches \textbf{77.9\%} case-level agreement with human pairwise preferences, dramatically outperforming CIDEr (47.5\%, near random), METEOR (55.7\%), and SODA\_c (60.7\%). At the model level, all four \metric variants attain Pearson $r{>}0.93$ with $p{<}0.025$, whereas the three traditional metrics yield $r{=}0.17$--$0.55$ with $p{>}0.3$ (not statistically significant). Combined with the cross-judge concordance ($W{=}0.925$) from Section~\ref{subsec:multi_judge}, this provides strong evidence that \metric is both \emph{judge-robust} and \emph{human-aligned}.

\subsection{Detailed Analysis of DP Merging}
\label{subsec:dp_merging}

This appendix expands on the headline result reported in Section~\ref{subsec:dp_merging_main} (Table~\ref{tab:dp_merging}), providing the full motivation, experimental setup, and additional safeguards.

\noindent\textbf{Motivation.}
A potential concern about \metric is that the DP-based segment-merging step might itself introduce evaluation flaws---for example, by hiding pathological over-segmentation behavior. We address this concern from two angles: (i)~we show that \emph{without} DP merging, fine-grained models that predict shorter segments are systematically and unfairly under-rated; (ii)~we show empirically that DP merging does \emph{not} introduce systematic bias, as evidenced by the human alignment study in Section~\ref{subsec:human_alignment}.

\begin{figure*}[htb]
    \centering
    \includegraphics[width=0.99\textwidth]{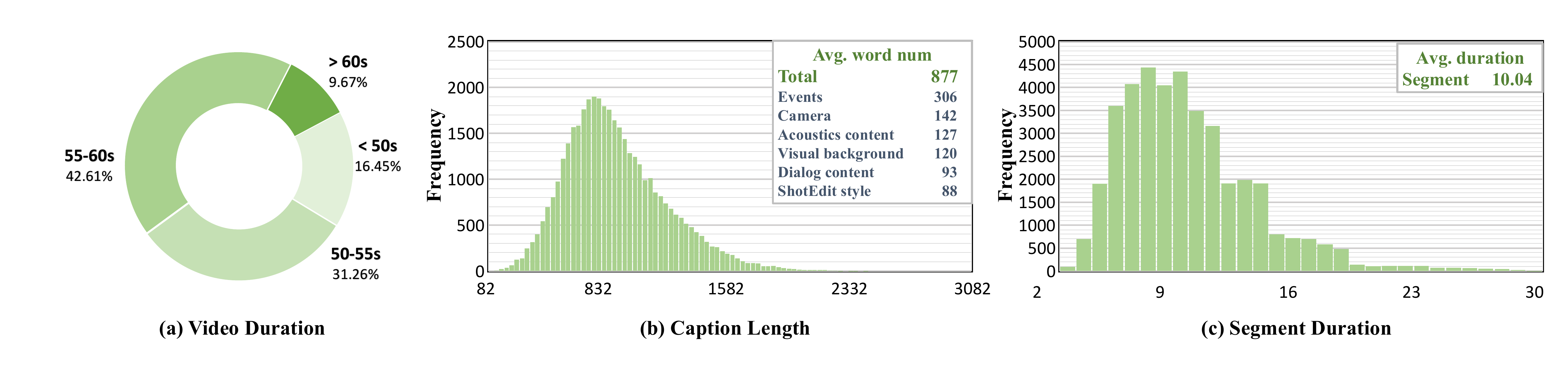}
    \caption{\textbf{Statistics of the training dataset \dataset.} \textbf{(a)} Video duration distribution; most videos (73.9\%) fall within 50-60 seconds. \textbf{(b)} Caption length distribution with per-dimension average word counts; annotations average 877 words per video across six dimensions. \textbf{(c)} Segment duration distribution; the average segment length is 10.04 seconds.}
    \label{fig:traindata_stat}
\end{figure*}

\noindent\textbf{Setup.}
Current MLLMs predict noticeably shorter segments than the ground truth (Figure~\ref{fig:traindata_stat}(c)): Qwen2.5-Omni-7B averages 7.0\,s and Qwen3-Omni-30B-A3B averages 4.2\,s, against a GT mean of 14.2\,s. We ablate \metric with and without DP merging on these two systems. As an external sanity check, we use the head-to-head human preference between them from the Elo study in Section~\ref{subsec:human_alignment}: humans prefer Qwen3-Omni in \textbf{75.0\%} of comparisons (75.0\% win / 16.7\% lose / 8.3\% tie).

\noindent\textbf{Findings.}
As Table~\ref{tab:dp_merging} (in main text Section~\ref{subsec:dp_merging_main}) shows, without DP merging the score gap shrinks to $+1.2$, severely mismatching the 75\% human preference for Qwen3-Omni. With DP merging, the gap widens to $+9.7$, matching human judgment. Two additional safeguards prevent metric gaming via degenerate over-segmentation:
\begin{itemize}[leftmargin=1.6em, nosep, topsep=2pt]
    \item Hard-matching \textbf{F1} and \textbf{mIoU} (reported in Table~\ref{tab:script_table}) directly penalize pathological over-segmentation: predicting many 1-second segments would cause F1/mIoU to collapse, which would be visible in the main results.
    \item The 77.9\% case-level agreement and Pearson $r{>}0.93$ in Section~\ref{subsec:human_alignment} are computed \emph{with} DP merging enabled, empirically verifying that DP merging does not introduce systematic bias against humans' preferred outputs.
\end{itemize}

\noindent\textbf{Summary.}
The three studies jointly show that \metric is (i)~robust across LLM judges (Kendall's $W{=}0.925$), (ii)~strongly aligned with human preferences (77.9\% case-level agreement, Pearson $r{>}0.93$), and (iii)~that the DP-merging step is necessary for fair evaluation rather than a source of bias.

\section{Additional Experimental Results}
\label{sec:additional_exp}

\noindent\textbf{Effect of Reward Weights.}
We investigate the sensitivity of our model to the reward weight coefficients $(\alpha_f, \alpha_l, \alpha_t, \alpha_c)$ in Equation~\ref{eq:reward}. As shown in Table~\ref{tab:reward_weights}, varying the weight of the coherence reward $R_C$ from 1.0 to 1.5 results in marginal performance differences across all metrics (less than 0.5\% on F1, mIoU, and \metric). This suggesting that the four reward components provide complementary supervision signals without requiring extensive hyperparameter tuning.

\begin{table}[h]
    \centering
    \begin{minipage}{0.48\textwidth}
        \centering
        \captionof{table}{Ablation study on reward weight coefficients.}
        \label{tab:reward_weights}
        \vspace{0.5em}
        \begin{tabular}{cccc}
        \toprule
        $(\alpha_f, \alpha_l, \alpha_t, \alpha_c)$ & \textbf{F1} & \textbf{mIoU} & \textbf{\metric} \\
        \midrule
        (0.5, 0.5, 1.0, 1.0) & 61.2 & 69.6 & \textbf{35.0} \\
        (0.5, 0.5, 1.0, 1.5) & 61.0 & 69.4 & 34.6 \\
        \bottomrule
        \end{tabular}
    \end{minipage}
    \hfill
    \begin{minipage}{0.48\textwidth}
        \centering
        \captionof{table}{Ablation study on SFT training epochs.}
        \label{tab:sft_epochs}
        \vspace{0.5em}
        \begin{tabular}{cccc}
        \toprule
        \textbf{SFT Training} & \textbf{F1} & \textbf{mIoU} & \textbf{\metric} \\
        \midrule
        Epoch 1 & 61.7 & 70.4 & 30.7 \\
        Epoch 2 & \textbf{62.4} & \textbf{70.7} & \textbf{32.6} \\
        \bottomrule
        \end{tabular}
    \end{minipage}
\end{table}

\noindent\textbf{Effect of SFT Training Epochs.}
We examine the impact of supervised fine-tuning (SFT) duration on model performance. As shown in Table~\ref{tab:sft_epochs}, extending SFT training from 1 to 2 epochs yields consistent improvements across all evaluation metrics. These results suggest that the \task task is inherently complex, requiring sufficient SFT training for the model to adequately learn the structured output format and multi-dimensional annotation capabilities. Moreover, a well-trained SFT checkpoint serves as a stronger initialization for the subsequent GRPO stage, enabling more effective reward-guided optimization. We therefore adopt two-epoch SFT training as our default configuration.

\begin{figure*}[t]
    \centering
    \includegraphics[width=0.99\textwidth]{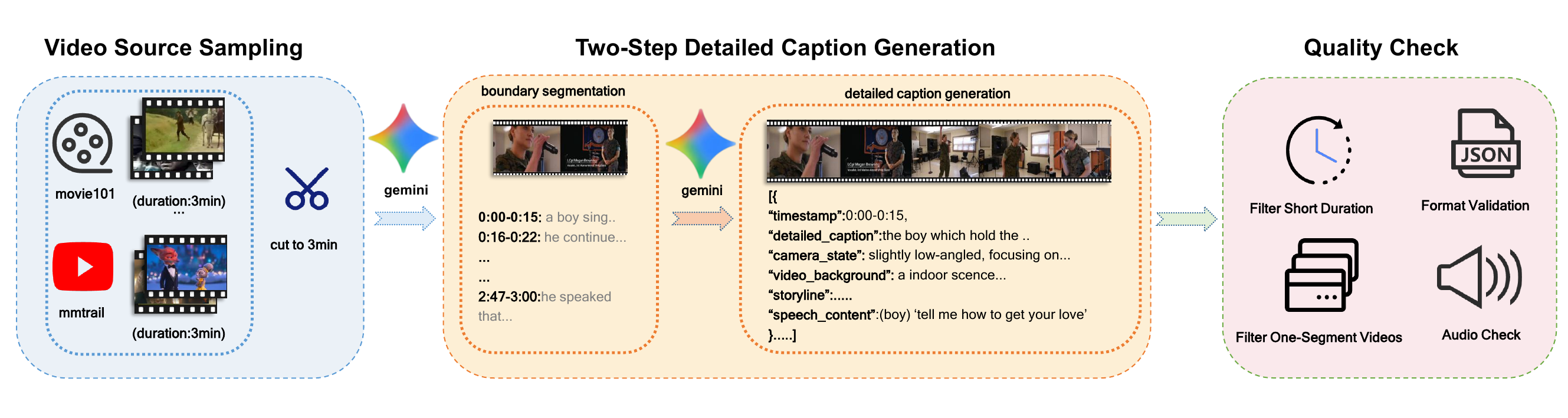}
    \caption{Overview of the synthetic training data construction pipeline for the training dataset \dataset.}
    \label{fig:data_pipeline}
\end{figure*}

\section{Details for Training Data and Benchmark Annotation}
\label{sec:data_details}

\subsection{Training Data Construction}
\label{subsec:training_data}
As illustrated in Figure~\ref{fig:data_pipeline}, we design a three-stage pipeline to synthesize high-quality training samples for the \task task. Detailed prompts are shown in Table~\ref{tab:stage1_prompt} and Table~\ref{tab:stage2_prompt}.

Figure~\ref{fig:traindata_stat} presents the detailed statistics of \dataset: \textbf{(a)} Video duration distribution, where the majority of videos (73.9\%) fall within the 50-60 second range. \textbf{(b)} Caption length distribution across dimensions, with annotations averaging 877 words per video spanning six dimensions. \textbf{(c)} Segment duration distribution, showing an average segment length of 10.04 seconds.

\subsection{Human Annotation Details for \benchmark}
\label{subsec:human_anno}

\begin{figure}[t]
    \centering
    \includegraphics[width=\columnwidth]{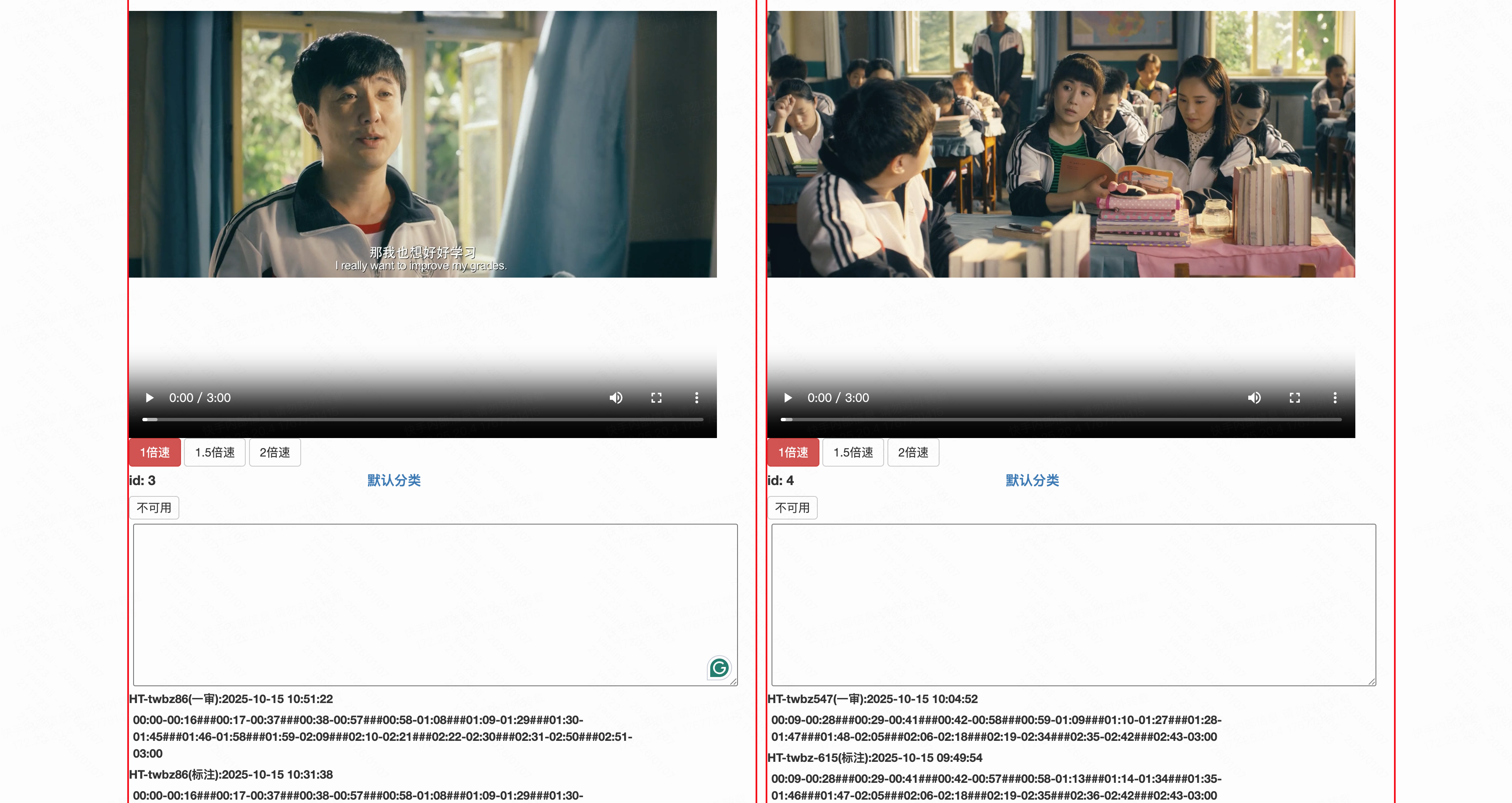}
    \caption{Interface page used for manual annotation during the construction of \benchmark.}
    \label{fig:human_anno_screenshot}
\end{figure}

The \benchmark is entirely annotated by human experts. We recruited approximately annotators through crowdsourcing platforms, and the complete annotation process spanned approximately one month. As shown in Figure~\ref{fig:human_anno_screenshot}, the annotation interface is designed to be intuitive and user-friendly, featuring clear instructions and real-time feedback mechanisms to facilitate efficient task completion. To ensure annotation quality, each sample was reviewed by at least one additional annotator.

\section{Implementation Details}
\label{sec:implementation}
Our training procedure consists of two stages: Supervised Fine-Tuning (SFT) and Reinforcement Learning via Group Relative Policy Optimization (GRPO). During the SFT stage, the model is fine-tuned for 2 epochs on 40K training samples, with a learning rate of 5e-5 and a global batch size of 128. In the subsequent GRPO phase, we utilize 2K training samples and employ a rollout size of 8 to compute group relative advantages. The learning rate, batch size, and number of epochs for RL alignment are set to 1e-5, 64, and 1, respectively. The KL penalty coefficient $\beta$ is set to 0.04. The weights for format, length, temporal alignment, and caption quality rewards are configured as 0.5, 0.5, 1.0, and 1.0. To facilitate long-video understanding, we set the maximum sequence length to 32K tokens. All videos are uniformly sampled at 2 Frames Per Second (FPS). We limit the maximum pixels per frame to 297,920 and the total pixels per video to 20,070,400, ensuring a balance between visual fidelity and computational efficiency. All experiments are conducted on 32x80G GPUs using DeepSpeed ZeRO-2.

\noindent\textbf{Baselines.}
To comprehensively evaluate the performance of our proposed method, we compare it against three categories of models. (1)~\textbf{Closed-source MLLMs}: leading commercial systems Gemini-2.5-Pro and Gemini-2.5-Flash~\cite{gemini}. (2)~\textbf{Open-source MLLMs}, further grouped by their primary design focus: general-purpose omni-modal models (Qwen2.5-Omni~\cite{qwen2d5omni}, Qwen3-Omni~\cite{Qwen3-Omni}, MiniCPM-o-2.6~\cite{minicpm}, video-SALMONN-2~\cite{videoSALMONN2}), which target unified multimodal understanding; human-centric models (HumanOmniV2~\cite{yang2025humanomniv2}), tailored to person-centered video reasoning; and domain-specialized captioners (ARC-Hunyuan-Video~\cite{ge2025arc}, UGC-VideoCaptioner~\cite{wu2025ugc}), optimized for caption-style generation on user-generated or short-form videos. (3)~\textbf{Expert Models} specialized for temporal video understanding: LongVALE~\cite{geng2025longvale}, TimeChat~\cite{timechat}, TimeSuite~\cite{timesuite}, and TimeExpert~\cite{timeexpert}. For each baseline, we use the official released checkpoints and follow the recommended inference configurations.

\section{Additional Qualitative Analysis}
\label{sec:qualitative}

We present qualitative comparisons among \model, Gemini-2.5-Pro, and Qwen-3-Omni on a representative sample from \benchmark, as illustrated in Figure~\ref{fig:case}. 

\model~achieves \textbf{fine-grained alignment} with the ground truth across all six annotation dimensions, as shown in the following:

\begin{itemize}[leftmargin=*,nosep]
    \item \textbf{Detailed Events:} Our model accurately identifies characters by their names (``Xia Luo'') and provides detailed appearance descriptions (e.g., ``a man with a distinctive, spiky haircut and a black suit''). The phrase ``continues to drive in circles'' demonstrates temporal awareness and scene continuity from preceding segments. Fine-grained actions such as ``leans forward to peer out the window, searching for something'' are faithfully captured.
    
    \item \textbf{Visual Background:} The model correctly recognizes the vehicle type (``luxury Maserati'') and simultaneously describes both interior and exterior environments (``the opulent courtyard seen through the windows''), maintaining spatial consistency.
    
    \item \textbf{Acoustics Content:} The model captures nuanced tonal transitions in speech (``Anxious, then commanding'') and identifies ambient sounds (``the hum of the car's engine'').
    
    \item \textbf{Dialogue Content:} Speaker attribution is precise, with each utterance correctly assigned to the corresponding character (``Xia Luo: ...''), and the conversational content aligns with the visual narrative.
    
    \item \textbf{Camera State:} Camera positioning (``inside the car, looking at the driver and passenger'') and shot transitions (``cuts to a close-up of the man in the back seat'') are accurately described.
    
    \item \textbf{Shot Editing Style:} The model provides purposeful analysis of editing choices, noting that ``interior shots focus on the characters' interactions and motivations.''
\end{itemize}

In contrast, Gemini-2.5-Pro~\cite{gemini} exhibits \textcolor{black}{severe hallucination} by misidentifying the male driver as ``a woman,'' fundamentally distorting the scene semantics. Qwen-3-Omni~\cite{Qwen3-Omni} \textit{misses the main event} entirely---instead of describing the conversation inside the car, it focuses on irrelevant background elements such as ``a man dressed in a red ceremonial uniform'' standing outside the building. These comparisons highlight \model's superior capability in accurate character recognition, consistent identity tracking across time, faithful event grounding, and comprehensive multi-dimensional annotation.


\section{Prompt Templates}
\label{sec:prompts}

We provide the detailed prompt templates used in our training data construction pipeline and evaluation framework.

\begin{table*}[ht]
    \centering
    \small
    
    \begin{tcolorbox}

\textbf{Stage-1 Prompt for Dense Timestamp Generation:}\\

You are given a video clip of around 3 minutes. Your task is to generate \textbf{dense captions} for the video. The goal is to segment the video into multiple meaningful intervals and provide detailed descriptions for each segment.

\vspace{0.5em}
\textbf{1. Segmentation Logic:}
\begin{itemize}[leftmargin=*,nosep]
    \item Split the video into natural segments according to \textbf{scene changes, shot transitions, events, character actions, or core content shifts}.
    \item Each minute usually contains \textbf{4--5 segments}, but prioritize the video's logic over strict numbers.
\end{itemize}

\vspace{0.5em}
\textbf{2. Caption Requirements:}
\begin{itemize}[leftmargin=*,nosep]
    \item For each segment, provide a \textbf{time range} in the format: \texttt{start\_time(0:00) - end\_time(0:05)}: caption
    \item The caption should be \textbf{concise but descriptive}, summarizing what happens in that segment.
    \item Include \textbf{characters, actions, objects, emotions, and scene details} where relevant.
    \item Avoid redundancy, but ensure that important visual and narrative information is captured.
\end{itemize}

\vspace{0.5em}
\textbf{3. Output Format:}
\begin{itemize}[leftmargin=*,nosep]
    \item A clean list of captions, each starting with the time range followed by the description.
    \item Timestamp format: \texttt{minutes:seconds} (e.g., \texttt{0:00 - 0:11}).
    \item Segment boundaries should be \textbf{clear and non-overlapping}---the end time of one segment and start time of the next should be at least 1 second apart.
    \item Example: if one segment ends at \texttt{0:11}, the next one should begin at \texttt{0:12} or later.
\end{itemize}

\vspace{0.5em}
\textbf{Output Example:}
\begin{verbatim}
**0:00 - 0:07**
A man walks into the dimly lit room and looks around cautiously.

**0:08 - 0:15**
He notices a woman sitting by the window, staring outside in silence.

**0:16 - 0:25**
The camera cuts to a close-up of his nervous expression.
\end{verbatim}

\end{tcolorbox}
\caption{The annotation prompt used in Stage-1 of training data construction. This prompt instructs Gemini-2.5-pro to segment videos into meaningful intervals with concise  captions.}
\label{tab:stage1_prompt}
\end{table*}

\begin{table*}[ht]
    \centering
    \small
    
    \begin{tcolorbox}

\textbf{Stage-2 Prompt for Multi-Dimensional Structural Caption Generation:}\\

You are given:
\begin{itemize}[leftmargin=*,nosep]
    \item A \textbf{video} (with both visual and audio content).
    \item A set of \textbf{ground-truth captions} (GT captions) with timestamps.
\end{itemize}

Your task is to \textbf{use the GT captions as rough references} for segmentation, but generate \textbf{richer and more detailed descriptions} directly from the video itself.

\vspace{0.5em}
\textbf{1. Segmentation:}
\begin{itemize}[leftmargin=*,nosep]
    \item Follow the \textbf{timestamps provided in the GT captions}. Each GT caption defines the rough boundaries of a segment.
    \item Timestamp format: \texttt{minutes:seconds} (e.g., \texttt{0:00 - 0:11}).
    \item Segment boundaries should be \textbf{clear and non-overlapping}---at least 1 second apart between consecutive segments.
\end{itemize}

\vspace{0.5em}
\textbf{2. Generation Logic:}\\
Do \textbf{not} simply extract or paraphrase the GT caption. Use the video itself to \textbf{expand with details}:
\begin{itemize}[leftmargin=*,nosep]
    \item \textbf{Characters}: actions, gestures, facial expressions, emotions.
    \item \textbf{Objects \& Setting}: relevant items, props, environment.
    \item \textbf{Camera}: framing, movement, zoom, transitions.
    \item \textbf{Storyline}: how the segment advances or changes the plot.
    \item \textbf{Speech}: actual dialogue attributed to speakers.
    \item \textbf{Acoustics}: speech tone, background music, sound effects.
    \item \textbf{Shooting Style}: special techniques (montage, flashback, dissolve, long take, etc.).
\end{itemize}

\vspace{0.5em}
\textbf{3. Output Format (JSON Schema):}
\begin{verbatim}
{
  "timestamp": "start_time - end_time",
  "segment_detail_caption": "Detailed description of what happens 
      (gestures, expressions, setting details, etc.).",
  "camera_state": "Camera angle, framing, zoom, and movement.",
  "video_background": "Setting, environment, or background elements.",
  "storyline": "How this segment fits into the larger narrative.",
  "shooting_style": "Long take, montage, flashback, intercut, 
      or special transition effects.",
  "speech_content": "Full character dialogues with speaker attribution.",
  "acoustics_content": "1) Tone of speech. 2) Background sounds or music."
}
\end{verbatim}

\end{tcolorbox}
\caption{The annotation prompt used in Stage-2 of training data construction. Given Stage-1 captions as segmentation references, we prompt Gemini-2.5-pro to generate more enriched multi-dimensional annotations by directly perceiving the video content, covering detailed events, camera state, background, storyline, shooting style, speech, and acoustics.}
\label{tab:stage2_prompt}
\end{table*}

\begin{table*}[ht]
    \centering
    \small
    
    \begin{tcolorbox}

\textbf{Judge Prompt for \metric Checklist Evaluation:}\\

You are a \textbf{strict evaluator} for fine-grained audio-enhanced video captions.

\vspace{0.3em}
You will receive:
\begin{enumerate}[leftmargin=*,nosep]
    \item A list of \textbf{ground-truth keypoints} already organized in 6 dimensions.
    \item One \textbf{model-generated caption} to evaluate.
\end{enumerate}

The ground-truth keypoints are already \textbf{atomic and accurate}. You only need to check whether each keypoint is \textbf{explicitly mentioned or clearly implied} in the model's caption.

\vspace{0.5em}
\textbf{Rules:}
\begin{itemize}[leftmargin=*,nosep]
    \item Mark a keypoint as correct if its meaning appears in the model's caption with the same or equivalent semantics.
    \item Ignore differences in phrasing, tense, or minor wording.
    \item Do NOT infer or guess beyond the caption content.
    \item Do NOT generate new keypoints or summaries.
    \item Do NOT output any text other than the required JSON.
\end{itemize}

\vspace{0.5em}
\textbf{Output Format (Strict JSON Only):}
\begin{verbatim}
{
  "by_dim": {
    "segment_detail_caption": {
        "correct_keypoints": [<string>, ...], "correct_count": <int>},
    "video_background": {
        "correct_keypoints": [<string>, ...], "correct_count": <int>},
    "acoustics_content": {
        "correct_keypoints": [<string>, ...], "correct_count": <int>},
    "shooting_style": {
        "correct_keypoints": [<string>, ...], "correct_count": <int>},
    "speech_content": {
        "correct_keypoints": [<string>, ...], "correct_count": <int>},
    "camera_state": {
        "correct_keypoints": [<string>, ...], "correct_count": <int>}
  }
}
\end{verbatim}

\vspace{0.5em}
\textbf{Input Template:}
\begin{verbatim}
Ground-truth keypoints (by dimension):
- segment_detail_caption: [<keypoint_1>, <keypoint_2>, ...]
- video_background:       [<keypoint_1>, <keypoint_2>, ...]
- acoustics_content:      [<keypoint_1>, <keypoint_2>, ...]
- shooting_style:         [<keypoint_1>, <keypoint_2>, ...]
- speech_content:         [<keypoint_1>, <keypoint_2>, ...]
- camera_state:           [<keypoint_1>, <keypoint_2>, ...]

Model-generated caption to evaluate:
<model_caption>
\end{verbatim}

\end{tcolorbox}
\caption{The judge prompt used for \metric using checklist score during evaluation. Given ground-truth keypoints decomposed into six dimensions and a model-generated caption, the judge model (Gemini-2-Flash) verifies whether each atomic keypoint is explicitly mentioned or semantically implied in the prediction, enabling fine-grained recall computation across all annotation dimensions.}
\label{tab:judge_prompt}
\end{table*}




\end{document}